\documentclass{article}
\usepackage{spconf,amsmath,amssymb,graphicx,booktabs,multirow}
\usepackage[table]{xcolor}

\usepackage{enumitem}
\setlist{nosep, leftmargin=14pt}

\usepackage{mwe} 

\graphicspath{{../plots/}{figures/}}

\def\x{{\mathbf x}}
\def\z{{\mathbf z}}
\def\L{{\cal L}}

\title{Federated training performs as well as centralized for 3D Latent Diffusion for Diagnosis-Conditioned Brain MRI Synthesis}
\name{Abhijith Shaji$^{1}$, Paul Kurian$^{1}$, Greg Ver Steeg$^{3}$, Paul M. Thompson$^{2}$, Jose-Luis Ambite$^{1}$}
\address{$^{1}$Information Sciences Institute, University of Southern California, Marina del Rey, CA, USA\\
$^{2}$Imaging Genetics Center, Mark and Mary Stevens Neuroimaging and Informatics Institute,\\
Keck School of Medicine, University of Southern California, Marina del Rey, CA, USA\\
$^{3}$University of California, Riverside, CA, USA\\
\texttt{ashaji@usc.edu}}

\begin{document}
%
\maketitle
\begin{abstract}
Generative models of brain MRI are useful to understand brain conditions, and to enrich scarce and imbalanced neuroimaging data. However, training these models requires large, multi-site training sets, which is challenging. Federated learning (FL) allows for training deep learning models without sharing data. We show that training a 3D conditional latent model federatedly (where data is kept private at each site) performs similarly to centralized training (where all the data have been copied to a single site). We evaluate three metrics: conditional fidelity (how well generated MRIs perform in classification), diversity (how the generated images match the real data distribution, using MultiScale-SSIM), and realism (how close generated images are to real images, using Fr\'{e}chet Inception Distance). 
Our results show that federated training is a viable route to develop multi-site generative models of brain MRI.



\end{abstract}
\section{Introduction}
\label{sec:intro}

Deep learning for neuro-imaging is limited by data availability and distribution. Cohorts are often small, disease classes are imbalanced, and scanner and protocol differences between sites hurt generalization. Generative models offer one remedy, by synthesizing realistic, labeled scans for augmentation, for balancing rare classes, or for sharing data without exposing patient data. Denoising diffusion probabilistic models (DDPM)~\cite{ho2020ddpm} now set the state of the art in image synthesis, and latent diffusion models (LDMs)~\cite{rombach2022ldm} make them tractable for 3D medical volumes by running the diffusion process in the compressed latent space of an autoencoder. 3D LDMs have been used to generate high-resolution brain MRIs to complement real ones\cite{pinaya2022brainldm}.

However, these models need large and diverse training data and brain MRI is held by separate hospitals and consortium, where data sharing is challenging due to privacy, patient consent, and data-governance constraints. Federated learning (FL)~\cite{mcmahan2017communication} allows training a deep learning model on multiple sites without sharing data. Each site is trained over its own private data, and only encrypted model parameters are exchanged and averaged (e.g., \cite{stripelis2024:patterns}). Diffusion models have been trained federatedly~\cite{peng2025feddiffusion}, and pretrained latent diffusion models have been adapted to heterogeneous clients in one-shot FL~\cite{chen2025fedbip}. FL has been used for discriminative neuroimaging tasks as well~\cite{stripelis2024:patterns}. 
Although latent diffusion has been proposed for privacy-preserving synthesis~\cite{shi2026privacyldm}, training a full 3D brain MRI LDM federatedly remains much less explored.

We answer this question with a controlled comparison of the same diagnosis-conditioned 3D LDM, with identical architecture, losses and sampling where one is trained federatedly across five naturally-distributed data sites and another centrally on the pooled data, and both are evaluated on the same held-out scans with paired random seeds. 

\section{Data}
\label{sec:data}
We use T1-weighted MRI from the three phases of Alzheimer's Disease Neuroimaging Initiative (ADNI)~\cite{veitch2019understanding} (ADNI-1, ADNI-2/GO and ADNI-3) and the National Alzheimer's Coordinating Center (NACC)~\cite{beekly2007nacc}, split into GE and Siemens/Philips scanner groups. Each of the five groups were represented as a federated site. Scans were reoriented, skull-stripped, bias-field corrected and affinely (9-DOF) registered to a common template at 2\,mm isotropic resolution, padded to $96\times112\times96$ voxels and min--max normalized to $[0,1]$. Each scan is labeled cognitively normal (CN) or Alzheimer's disease (AD).

Table~\ref{tab:data} summarizes the splits. Training, Validation and Test sets are subject disjoint. 
%
ADNI subjects contribute several longitudinal scans, while NACC is mostly one scan per subject. 


\begin{table}[t]
\centering
\caption{Scans per site and diagnosis (CN/AD). Number of Subjects in parentheses.}
\label{tab:data}
\setlength{\tabcolsep}{3.5pt}
\footnotesize
\begin{tabular}{lccc}
\toprule
Site & Train & Validation & Test \\
\midrule
ADNI-1      & 807/741 (436)  & 76/113 (59)   & 88/127 (61) \\
ADNI-2/GO   & 1120/625 (593) & 74/52 (48)    & 57/64 (39) \\
ADNI-3      & 847/188$^\dagger$ (665) & -- & -- \\
NACC GE     & 978/761 (1530) & 101/95 (165)  & 102/107 (180) \\
NACC Si/Ph$^*$  & 891/363 (1048) & 137/49 (157)  & 131/34 (142) \\
\midrule
Total       & 4643/2678 (4028) & 388/309 (418) & 378/332 (417) \\
\bottomrule
\end{tabular}\\[2pt]
{\scriptsize $^\dagger$plus 16 scans without diagnosis. $^*$Si/Ph: Siemens/Philips.}
\end{table}

\section{Methods}
\label{sec:methods}

\subsection{3D autoencoder}
\label{ssec:ae}

The autoencoder is a 3D AutoencoderKL from MONAI Generative Models~\cite{pinaya2022brainldm,pinaya2023generative,cardoso2022monai} with channels $(64,128,256)$, two residual blocks per level, GroupNorm (16 groups) and self-attention at the lowest resolution. The encoder maps a volume $\x$ to a Gaussian posterior with mean $\boldsymbol{\mu}$ and standard deviation $\boldsymbol{\sigma}$ over an $8\times24\times28\times24$ latent space and the decoder reconstructs $\x$ from a latent sample $\z=\boldsymbol{\mu}+\boldsymbol{\sigma}\odot\boldsymbol{\epsilon}$. It is trained with
\begin{equation}
\L_{\text{AE}} = \L_{1}^{\text{w}} + 10^{-6}\,\L_{\text{KL}} + 0.01\,\L_{\text{perc}} + 0.01\,\L_{\text{adv}},
\label{eq:ae}
\end{equation}
where $\L_{1}^{\text{w}}$ is weighted L1 loss with brain voxels ($x>0.05$) weighted $\times10$, $\L_{\text{perc}}$ is perceptual loss computed on 2D slices (20\%) with a SqueezeNet backbone\cite{iandola2016squeezenet}, and $\L_{\text{adv}}$ is least-squares adversarial loss from a 3D PatchGAN discriminator, enabled after five warm-up epochs.


\subsection{Diagnosis-conditioned latent diffusion}
\label{ssec:ldm}
Following the original LDM design~\cite{rombach2022ldm}, the diffusion model is trained on sampled latents $\z$ of the frozen autoencoder, multiplied by a scale factor $s=1/\mathrm{std}(\z)$ so that their variance matches the unit-variance diffusion noise. The denoiser is a 3D DiffusionModelUNet~\cite{pinaya2023monaigen} with channels $(128,256,512)$, two residual blocks per level, and cross-attention at the two lower resolutions. The diagnosis $c\in\{\text{CN},\text{AD},\varnothing\}$ is mapped by a learnable embedding table to a 64-dimensional token consumed by cross-attention. We use a DDPM~\cite{ho2020ddpm} forward process with $T=1000$ steps in which noise is added gradually according to a variance schedule $\beta_1,\dots,\beta_T$ (scaled-linear from $\beta_1=0.0015$ to $\beta_T=0.0195$). $\bar\alpha_t=\prod_{i=1}^{t}(1-\beta_i)$ is the fraction of the signal that remains after $t$ steps. The UNet $\boldsymbol{\epsilon}_\theta$ is trained to predict the added noise by minimizing
\begin{equation}
\L_{\text{LDM}} = \mathbb{E}_{\z,\boldsymbol{\epsilon},t,c}\left\| \boldsymbol{\epsilon} - \boldsymbol{\epsilon}_\theta\big(\sqrt{\bar\alpha_t}\, s\z + \sqrt{1-\bar\alpha_t}\,\boldsymbol{\epsilon},\, t,\, c\big) \right\|_2^2 .
\end{equation}
where $\boldsymbol{\epsilon}\sim\mathcal{N}(\mathbf{0},\mathbf{I})$ is Gaussian noise of the latent's size, $t$ is drawn uniformly from $\{1,\dots,T\}$.
For classifier-free guidance~\cite{ho2022cfg}, the diagnosis is replaced by the null token $\varnothing$ with probability 0.15, so a single network learns both the conditional and the unconditional model. At sampling time, the two predictions are combined as
\begin{equation}
\hat{\boldsymbol{\epsilon}} = \boldsymbol{\epsilon}_\theta(\z_t,t,\varnothing) + w\,\big(\boldsymbol{\epsilon}_\theta(\z_t,t,c)-\boldsymbol{\epsilon}_\theta(\z_t,t,\varnothing)\big),
\end{equation}
where the guidance scale $w=1$ gives conditional and $w>1$ strongly guided samples. We sample with Denoising Diffusion Implicit Model (DDIM)~\cite{song2021ddim} (200 steps), decode $\z_0/s$ with the autoencoder and clip the output to $[0,1]$.

\subsection{Federated and centralized training}
\label{ssec:fl}
In both settings, the pipeline is trained in two consecutive stages. \emph{Stage 1} trains the autoencoder (Sec.~\ref{ssec:ae}) to compress each scan into a compact latent and reconstruct it and \emph{Stage 2} trains the diffusion UNet (Sec.~\ref{ssec:ldm}).

\textbf{Federated.} Training runs on the SPHERE research testbed \cite{mirkovic2026sphere} with the MetisFL federated learning framework \cite{stripelis2023:metisfl,stripelis2024:patterns} over five sites with the data in Table~\ref{tab:data}. Each site has two NVIDIA L40S GPUs. In each round every learner trains the community model for one local epoch on its own data, and the server aggregates the learners' parameters by federated averaging, weighting each learner by its number of training scans. All five learners take part in every round, and each trains data-parallel on its two GPUs. Scans, latents and labels never leave a site. Sites share the model parameters each round (in stage 1, the autoencoder's optimizer moments are also shared). During stage 1 the discriminator is kept local to each site and never shared, since only the autoencoder is needed downstream.

\textbf{Centralized.} The same models are trained on the pooled training set with distributed data-parallel training on 16 NVIDIA RTX A6000 GPUs (4 nodes). This is the upper bound that federated training is measured against and it requires all scans in one place.

\textbf{Scale factor.} The scale factor $s$ requires the standard deviation of the latents over all training scans. Each site encodes its scans with the frozen federated autoencoder and shares only three numbers: the count $N_k$, the sum $S_k=\sum z$ and the sum of squares $Q_k=\sum z^2$ over all latent values. The server computes $N=\sum_k N_k$, $\mathrm{mean}=\sum_k S_k/N$, $\mathrm{var}=\sum_k Q_k/N-\mathrm{mean}^2$ and $s=1/\sqrt{\mathrm{var}}$, which equals the value computed on the pooled latents exactly. This yields $s=0.9977$ for the federated autoencoder (per-site standard deviations 1.0007--1.0053) and $s=1.0035$ for the centralized one.

\textbf{Training details.} All models are trained in full fp32 precision with Adam. Stage 1 uses a learning rate of $10^{-4}$ and a batch of 2 per GPU in both settings. In this stage, checkpoints are selected by validation PSNR within the first 500 epochs or rounds (centralized epoch 400, federated round 300). Stage 2 uses a batch of 10 per GPU, a learning rate of $5\cdot10^{-5}$ with a linear warm-up, then constant, and gradient-norm clipping at 1.0. We evaluate both the centralized and federated models after 2000 epochs/rounds.


\subsection{Evaluation}
\label{ssec:eval}
We generate 1,500 CN and 1,500 AD volumes at $w=1$ (DDIM, 200 steps) using both the central and federated LDMs. Synthetic sets are compared with the pooled real test and validation scans (766 CN, 641 AD). To score of the generators, we also evaluate one random draw of 1,500 real training scans per class in exactly the same way.
\begin{itemize}
  \item \emph{Conditioning fidelity (AD vs.\ CN).} We train a 3D DenseNet-121~\cite{huang2017densely} AD vs CN classifier on the 7,321 labeled real training scans for 100 epochs with early stopping (patience=10) to prevent overfitting\cite{prechelt2002early}, using class-weighted cross-entropy and AdamW optimizer \cite{loshchilov2017decoupled}. We report the AUC for each classifier.
  \item \emph{Diversity.} Mean MS-SSIM~\cite{wang2003multiscale} computed with MONAI's \texttt{MultiScaleSSIMMetric}. We used 5 scales with the default weights, Gaussian kernel of size 4 and $\sigma=1.5$, data range 1. MS-SSIM is computed for every pair of volumes within a set (each unordered pair $i<j$ once, no random subsampling), separately for each class, and averaged. Higher values mean more similar samples , i.e.\ lower diversity.
  \item \emph{Realism.} Fréchet Inception Distance (FID)~\cite{heusel2017fid} on 2048-dimensional features of a 3D ResNet-50 pretrained on 23 medical datasets (MedicalNet~\cite{chen2019med3d}). The reference is always the pooled real test and validation scans. Lower FID means that the feature distribution of the synthetic volumes is closer to that of reference (real) scans. For computing FID, we used a brain mask for the real and synthetic images. Voxels above 0.01 after min--max scaling, of which only the largest 26-connected component is kept and its holes filled. Voxels outside the mask are set to 0 and those inside are z-scored, the intensity normalization MedicalNet was trained with.

\end{itemize}

\section{Results}
\label{sec:results}

\subsection{Reconstruction}
\label{ssec:res_ae}

Table~\ref{tab:ae} compares the selected autoencoders. Pooled over the four validation sites, the federated autoencoder performs similar to the centralized autoencoder with PSNR (26.09 vs.\ 26.30\,dB) and equal SSIM. The ADNI sites are on par and the small gap is concentrated at the two NACC sites. 

\begin{table}[t]
\centering
\caption{Autoencoder reconstruction on the validation set (mean over scans).}
\label{tab:ae}
\setlength{\tabcolsep}{3pt}
\footnotesize
\begin{tabular}{lcccccc}
\toprule
 & \multicolumn{2}{c}{PSNR (dB) $\uparrow$} & \multicolumn{2}{c}{SSIM $\uparrow$} & \multicolumn{2}{c}{MAE $\downarrow$} \\
Site ($n$) & Fed & Cen & Fed & Cen & Fed & Cen \\
\midrule
ADNI-1 (189)     & 25.51 & 25.48 & 0.981 & 0.978 & 0.0384 & 0.0386 \\
ADNI-2/GO (126)  & 26.19 & 26.30 & 0.982 & 0.978 & 0.0352 & 0.0352 \\
NACC GE (196)    & 26.19 & 26.55 & 0.973 & 0.976 & 0.0351 & 0.0341 \\
NACC Si/Ph (186) & 26.51 & 26.84 & 0.971 & 0.972 & 0.0341 & 0.0334 \\
\midrule
Pooled (697)     & 26.09 & 26.30 & 0.976 & 0.975 & 0.0357 & 0.0353 \\
\bottomrule
\end{tabular}
\end{table}

\subsection{Generation}
\label{ssec:res_gen}

Table~\ref{tab:gen} compares federated and centralized training over the three metrics.

\textbf{Conditioning fidelity is slightly better for federated samples.} The real-data classifier recognizes the intended diagnosis of synthetic scans as reliably as that of real patients: AUC 0.934 for federated and 0.922 for centralized generated samples, vs.\ 0.920 for real test and validation scans. The federated--centralized AUC difference is small. Balanced accuracy (0.873 vs.\ 0.867), sensitivity (0.803 vs.\ 0.791) and specificity (0.943 in both) do not differ significantly. Both models produce CN samples that are more clearly CN than real CN scans (specificity 0.943 vs.\ 0.854) and AD samples that are somewhat less clearly AD than real AD scans (sensitivity 0.80 and 0.79 vs.\ 0.86).

\textbf{Federated samples are more realistic.} Federated samples are closer to real scans than centralized ones for both classes: (FID $\times10^{-3}$): 1.25 vs.\ 1.89 for CN, 1.11 vs.\ 1.88 for AD and 1.14 vs.\ 1.84 pooled. Both remain clearly distinguishable from real data, whose training draw scores FID $0.0156\cdot10^{-3}$ against the held-out scans. In both models, 90--95\% of FID comes from the mean-shift term $\|\mu_1-\mu_2\|^2$, compared with 8--17\% for the real training scans, whose small FID comes mostly from the covariance term. The synthetic sets thus vary about as much as real scans but are offset from them as a whole, which points to a systematic difference shared by both pipelines, plausibly the autoencoder's reconstruction, rather than to the training paradigm.

\textbf{Diversity is lower for federated samples.} Centralized samples are nearly as diverse as real scans (MS-SSIM 0.905 and 0.900 for CN and AD, vs.\ 0.896 and 0.890 for both the real training draw and the real test and validation scans), whereas federated samples are more alike (0.918 and 0.915).

\begin{table}[t]
\centering
\caption{Generated samples ($w=1$, 1,500 per class and model) compared with real test+validation scans.}
\label{tab:gen}
\setlength{\tabcolsep}{2.6pt}
\footnotesize
\begin{tabular}{llccc}
\toprule
Class & Set & FID{\scriptsize($\times10^{-3}$)} & MS-SSIM & AUC \\
\midrule
\multirow{4}{*}{CN}     & Cen            & 1.889  & 0.905     & --    \\
                        & Fed            & 1.254  & 0.918 & --    \\
                        & Real train     & 0.0284 & 0.896     & --    \\
                        & Real test+val  & --    & 0.896     & --    \\
\midrule
\multirow{4}{*}{AD}     & Cen            & 1.878  & 0.900     & --    \\
                        & Fed            & 1.109  & 0.915 & --    \\
                        & Real train     & 0.0331 & 0.890   & --    \\
                        & Real test+val  & --    & 0.890     & --    \\
\midrule
\multirow{4}{*}{Pooled} & Cen            & 1.843  & --        & 0.922 \\
                        & Fed            & 1.143  & --        & 0.934 \\
                        & Real train     & 0.0156 & --        & --    \\
                        & Real test+val  & --    & --        & 0.920 \\
\bottomrule
\end{tabular}
\end{table}

\section{Discussion}
\label{sec:discussion}

\textbf{Federated training is a viable alternative.} Across reconstruction, conditioning fidelity and realism, the federated pipeline matches or exceeds the centralized reference with 99\% of the autoencoder PSNR, an AD/CN classification of its samples at least as accurate as for centralized samples and for real patients, and lower FID for both diagnoses. However, federated samples cluster more tightly around typical anatomy and reproduce fewer of the atypical brains present in real data. A generator that concentrates on typical anatomy lands closer to the bulk of the real distribution, which lowers distribution distances dominated by the mean shift. A plausible cause is that averaging updates from heterogeneous sites pulls the model towards anatomy shared by all sites, the federated autoencoder may contribute as well. 
The diversity gap is moderate (0.918 vs.\ 0.905, with 0.896 for real data) and not a collapse with federated samples still carrying their diagnoses.

\textbf{Privacy as the deciding factor.} These small differences have to be weighed against what each setting requires. Centralized training needs every scan copied to one place, which multi-site clinical data rarely permits. In the federated setting, scans, latents and labels never leave their site. The only shared quantities are model parameters and, once, three summary statistics per site for the scale factor.



In conclusion, we trained the same diagnosis-conditioned 3D latent diffusion model for brain MRI federatedly across five sites and centrally on pooled data, and compared both on held-out test scans across three metrics: conditional fidelity, diversity, and realism. Federated training matched the centralized reference in autoencoder reconstruction (99\% of PSNR) and in how reliably samples express the intended diagnosis, and produced samples closer to real scans in MedicalNet feature space, at the cost of moderately lower sample diversity. Since it achieves this without any scan leaving its site, federated training is a practical route to generative models built on multi-site clinical data that cannot be pooled.

\section{Compliance with ethical standards}
\label{sec:ethics}

This retrospective analysis used de-identified data obtained from ADNI and NACC. All studies were approved by the institutional review boards of the participating sites, and participants provided written informed consent. The analysis used data under the applicable cohort data use agreements.

\section{Acknowledgments}
\label{sec:acknowledgments}
This work was supported in part by NIH grants RF1AG081571, U01AG068057, R01MH131806, and S10OD032285. We thank Dr.\ Jelena Mirkovic for providing GPU computing resources on the SPHERE testbed.

\bibliographystyle{IEEEbib}
\bibliography{refs}

\end{document}